\documentclass[10pt,journal,compsoc]{IEEEtran}

\usepackage{amsmath}
\usepackage{amsfonts}
\usepackage{amssymb}

\usepackage{graphicx}
\usepackage{booktabs}
\usepackage[flushleft]{threeparttable}
\usepackage{multirow}
\usepackage{array}
\usepackage{tabularx}

\usepackage{algorithm}
\usepackage{algorithmic}

\usepackage[
    caption=false,
    font=footnotesize,
    labelfont=sf,
    textfont=sf
]{subfig}

\usepackage{microtype}

\usepackage{cite}
\usepackage{url}
\usepackage[hidelinks]{hyperref}
\usepackage[numbers,sort&compress]{natbib}
\usepackage{xspace}
\usepackage{enumitem}

\usepackage{stfloats}
\usepackage{placeins} 
\usepackage{balance}

\usepackage{xcolor}

\graphicspath{{pictures/}}

\begin{document}

\setcounter{secnumdepth}{3}

\title{
Provisioning Robust and Efficient 
\\
LiDAR Perception for Autonomous Driving
}

\author{
Tingxu~Han$^*$,
Zijie~Shen$^*$,
An~Guo,
Lyuke Wang,
Chunrong~Fang,
Zhenyu~Chen,
and~Wei~Song%
\thanks{
Tingxu Han, Zijie Shen, Chunrong Fang, and Zhenyu Chen are with the
State Key Laboratory for Novel Software Technology,
Nanjing University, Nanjing 210093, China.
Tingxu Han and Zijie Shen contribute equally.
(e-mail: txhan@smail.nju.edu.cn;
231820004@smail.nju.edu.cn;
fangchunrong@nju.edu.cn;
zychen@nju.edu.cn)
}
\thanks{
An Guo is with the Department of Computing, Hong Kong Polytechnic University,
Hong Kong SAR, China
(e-mail: an.guo@polyu.edu.hk).
}
\thanks{Lyuke Wang is with the School of Science and Engineering, The Chinese University of Hong Kong, Shenzhen, China (e-mail: 223010161@link.cuhk.edu.cn).}

\thanks{
W. Song is with the Information and Communication Technology School,
Griffith University, Gold Coast, QLD, 4215, Australia
(e-mail: w.song@griffith.edu.au).
}
}

\maketitle

\begin{abstract}
LiDAR-based 3D object detection is a core perception service for autonomous driving, yet practical deployment requires maintaining robustness to point-cloud corruption while meeting latency and memory requirements.
Existing approaches typically address robustness and resource efficiency separately, resulting in compact detectors that are fragile under corruption or robust detectors that remain costly to deploy.
We therefore formulate LiDAR perception provisioning as constructing a detector that remains accurate under point-cloud corruptions while meeting constraints on inference latency and peak GPU memory.
To achieve robustness, we first craft corrupted samples that mimic extreme weather conditions.
Then we train candidate detectors offline by ordering clean and corrupted samples by perturbation sensitivity, through which detectors learn more from hard corrupted samples.
We estimate perturbation sensitivity using an ensemble of seven heterogeneous 3D detectors by tracking changes in confidence, BEV overlap, and unmatched detections.
To improve efficiency, we narrow the network and use offline neighborhood-density measurements to determine how many points and neighbors each sampling stage processes.

We evaluate our method using AP (Average Precision) under clean and corrupted conditions, inference latency, and peak allocated GPU memory.
On KITTI for Car detection with PointRCNN, our method improves Moderate AP under corruption by 3.32 points over the original detector while reducing mean latency by 19.8\% and peak allocated GPU memory by 53.9\%. 
Across two datasets, two detector architectures, and three object classes, our method improves AP under corruption by 4.03 points on average over the corresponding original detectors, while reducing mean inference latency and peak allocated GPU memory by 14.4\% and 38.0\%, respectively. 
Our code is available at \url{https://github.com/drunklibai/robust-efficient-lidar-perception}.

\end{abstract}

\begin{IEEEkeywords}
Service provisioning,
quality of service,
LiDAR perception service,
corruption robustness,
resource efficiency.
\end{IEEEkeywords}

\section{Introduction}
\label{sec:intro}

\IEEEPARstart{L}{iDAR-based} 3D object detection is a core perception service for autonomous driving.
It converts streaming point clouds into the locations, dimensions, and orientations of surrounding objects, and serves downstream consumers such as tracking, prediction, and motion planning that issue per-frame inference requests and expect detections within a response-time budget~\cite{shi2019pointrcnn,yan2018second,shi2020pvrcnn}.
Onboard perception must process every incoming frame within its latency budget and available accelerator memory while remaining reliable under degraded input.~\cite{jiang2025energy,kong2023robo3d}.
Extreme weather such as rain and snow, measurement noise, and point-density variations can corrupt LiDAR measurements and substantially degrade 3D detection performance~\cite{kong2023robo3d,dong2023benchmarking}.
A deployable LiDAR perception service must therefore satisfy two coupled requirements: robustness to point-cloud corruption under extreme weather conditions such as rain, snow, noise, and point-density variation, and resource efficiency in terms of inference latency and memory use.

Existing research largely pursues these two objectives in isolation. 
Robustness-oriented approaches employ data augmentation, domain generalization or adaptation, and corruption-aware training to improve 3D detection under distribution shifts and adverse conditions~\cite{lehner2022vfield,li2023gpa3d,kong2023robo3d}.
Most of these methods modify only the training process, leaving the detector architecture and point-sampling workload unchanged. As a result, they improve robustness without reducing inference latency or memory use. 
Efficiency-oriented approaches instead rely on weight or channel pruning, knowledge distillation, lightweight detector design, or computation-aware point sampling~\cite{han2015learning,liu2017learning,hinton2015distilling,yang2020_3dssd,zhang2022iassd}.
Their reduction decisions are typically guided by parameter magnitudes, channel statistics, or accuracy on clean data, with limited consideration of detector behavior under point-cloud corruption.

To address this gap, we cast the problem as offline provisioning of a LiDAR perception service: before deployment, a provider constructs a fixed detector that meets both requirements, so that no adaptation is needed while serving requests. We develop a provisioning framework that builds such a detector to improve robustness to point-cloud corruption while reducing inference latency and GPU memory use.
For robustness, we first pair each clean training frame with a corrupted counterpart. An ensemble of seven 3D detectors then measures perturbation sensitivity from prediction discrepancies between each clean--corrupted pair, and we order all clean and corrupted samples from low to high sensitivity for training. 
For efficiency, we narrow the detector network and, for PointRCNN, use offline point-density statistics to reduce the numbers of points and neighbors processed at each sampling stage. The resulting detector is trained offline and then deployed for inference.

Experiments on KITTI~\cite{geiger2012kitti} and ONCE~\cite{mao2021once} using PointRCNN and SECOND show that our framework improves robustness under corruption while reducing inference cost. 
The main contributions are summarized as follows:
\begin{itemize}
    \item We establish an offline provisioning setting for LiDAR perception services that considers robustness to point-cloud corruptions and deployment efficiency. 

    \item We develop a provisioning framework that integrates two coordinated components. Specifically, we rank clean--corrupted pairs by perturbation sensitivity for robust training. We reduce inference cost through channel-width reduction and data-informed budget allocation.

    \item We evaluate the framework on KITTI and ONCE using the point-based PointRCNN and voxel-based SECOND across three object classes. Compared with the corresponding original detectors, the provisioned detectors improve AP under corruption by 4.03 points on average while reducing mean inference latency and peak allocated GPU memory by 14.4\% and 38.0\%, respectively.
\end{itemize}

\section{Related Work}
\label{sec:related}

\subsection{LiDAR Detection under Point-Cloud Corruptions}

Research on robust LiDAR detection mainly follows two directions: common-corruption benchmarking and weather-specific robustness enhancement.
\citet{dong2023benchmarking} and \citet{kong2023robo3d} introduce KITTI-C, nuScenes-C, and Waymo-C to evaluate 3D detectors under realistic point-cloud corruptions, while Robo3D covers adverse weather, external disturbances, and sensor failures.
Weather-specific methods synthesize physically grounded fog and snowfall on clean LiDAR scans and retrain detectors under the corresponding conditions~\cite{hahner2021fog,hahner2022snowfall}.
These studies primarily aim to evaluate or improve detection robustness, without simultaneously reducing inference latency and memory use.
Our offline provisioning framework instead combines perturbation-sensitivity-based data ordering with resource-efficient detector construction.

\subsection{Data Prioritization for Robust Training}

Curriculum learning presents training examples in a meaningful progression, typically from easy to hard, whereas self-paced learning determines this progression from the learner's current loss~\cite{bengio2009curriculum,kumar2010selfpaced}.
Curricular Object Manipulation applies this idea to LiDAR detection by dynamically reweighting object-level losses and progressively inserting harder objects through ground-truth augmentation~\cite{zhu2023curricular}.
Other data-centric methods prioritize selected training examples: core-set active learning chooses representative samples for annotation, while online hard example mining retains high-loss regions during each update~\cite{sener2018coreset,shrivastava2016ohem}.
Our method instead changes only the training order.
It retains all clean and corrupted frames without changing their loss weights, assigns each clean--corrupted pair a fixed perturbation-sensitivity score based on the prediction discrepancy of an ensemble of pretrained reference detectors, and presents the complete mixed set from low to high perturbation-sensitivity.

\subsection{Efficient LiDAR Detection and Point Processing}

Efficient LiDAR detection has been studied through representation design, point processing, and model compression.
SECOND uses sparse convolutions to avoid computation on empty voxels, while PointPillars organizes points into vertical pillars for efficient 2D convolution~\cite{yan2018second,lang2019pointpillars}.
For point-based detection, PointNet++-based detectors such as PointRCNN progressively downsample points and aggregate a bounded number of local neighbors~\cite{qi2017pointnetpp,shi2019pointrcnn}.
IA-SSD further improves efficiency by learning which foreground points to retain during downsampling~\cite{zhang2022iassd}.
Model-compression methods reduce network size through channel or structural pruning and knowledge distillation~\cite{liu2017learning,fang2023depgraph,hinton2015distilling}.
PointDistiller adapts structured knowledge distillation to point- and voxel-based 3D detectors~\cite{zhang2023pointdistiller}.
These approaches reduce cost through representations, learned sampling, pruning, or distillation, but do not combine their efficiency mechanisms with training driven by detector behavior under point-cloud corruption.
Our offline provisioning framework instead narrows network widths and, for PointRCNN, uses offline neighborhood-density statistics to set fixed point and neighbor budgets for each sampling stage; for SECOND, we use offline data statistics to set a fixed voxel budget.
It requires neither a learned sampler nor teacher--student distillation; ensemble predictions are used only to compute the perturbation-sensitivity scores that determine training order.

\section{Robust and Resource-Efficient LiDAR Detector Construction}
\label{sec:method}

\begin{figure*}[!t]
  \centering
  \includegraphics[width=\textwidth]{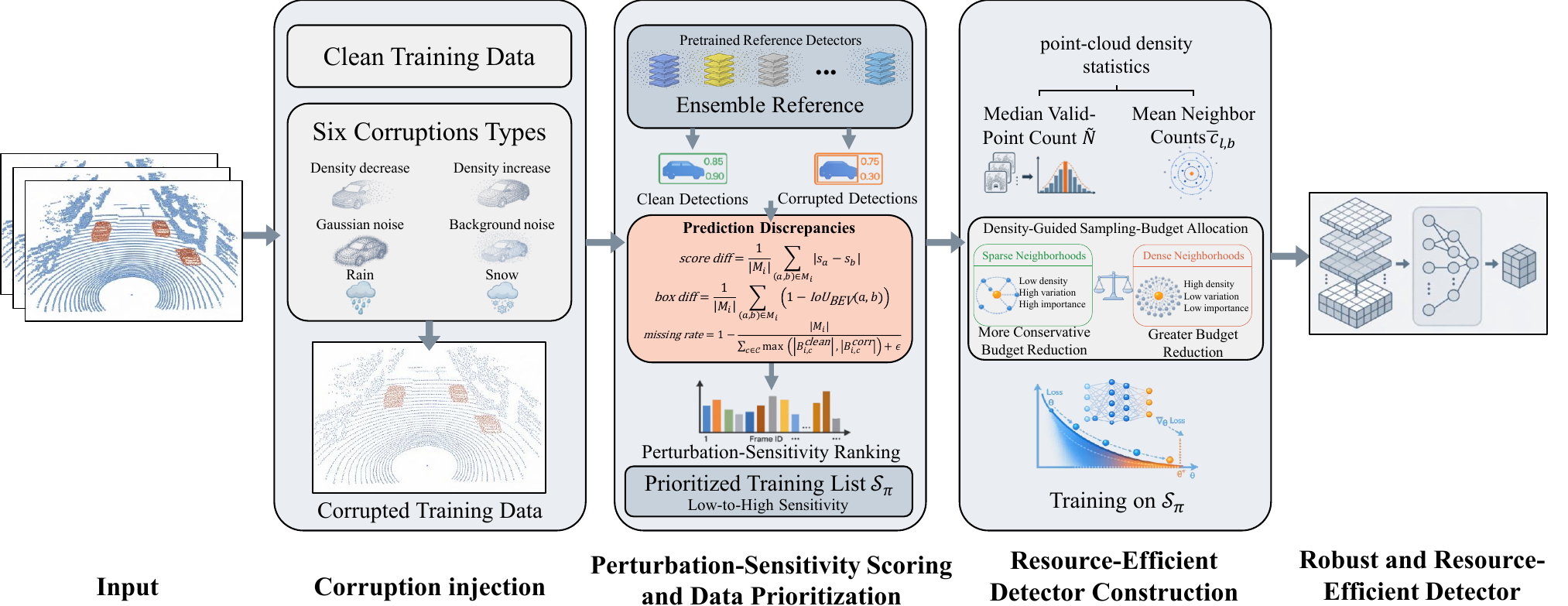}
  \caption{Offline profile-construction pipeline. Paired clean and corrupted point clouds are scored by an ensemble reference for clean-versus-corrupted prediction inconsistency. The reordered training data are combined with a density-guided sampling-budget allocation to train a compact detector.}
  \label{fig:methodology_pipeline}
\end{figure*}

\subsection{Problem Setting and Framework Overview}
A deployable LiDAR perception for autonomous driving must remain accurate under point-cloud corruptions while limiting inference latency and GPU memory use. We therefore construct a fixed detector offline to preserve clean-data average precision (AP), improve AP under corruption, and reduce mean inference latency and peak allocated GPU memory relative to the original detector.

Let $\mathcal{D}=\{(\mathbf{x}_i,\mathbf{y}_i)\}_{i=1}^{N}$ denote the clean training set, where $\mathbf{x}_i$ is a LiDAR point cloud and $\mathbf{y}_i$ contains its 3D object annotations. Let $\mathcal{A}$ denote the target detector architecture and $f_{\theta_0}$ the corresponding original detector with its default configuration. Given $\mathcal{D}$, a corruption operator $C(\cdot)$, $\mathcal{A}$, and $K$ pretrained reference detectors $\{f_k\}_{k=1}^{K}$, our framework constructs a robust and resource-efficient detector $f_{\theta'}$.

For any detector $f$, let $\mathrm{AP}{\mathrm{cln}}(f)$ and $\mathrm{AP}{\mathrm{cor}}(f)$ denote its AP on clean and corrupted data, and let $\operatorname{Lat}(f)$ and $\operatorname{Mem}(f)$ denote its mean inference latency and peak allocated GPU memory. The construction has two objectives:
\begin{equation}
\begin{aligned}
\text{maximize}\quad &
\{\mathrm{AP}_{\mathrm{cln}}(f),
      \mathrm{AP}_{\mathrm{cor}}(f)\}, \\
\text{minimize}\quad &
\{\operatorname{Lat}(f),
      \operatorname{Mem}(f)\}.
\end{aligned}
\label{eq:construction_objectives}
\end{equation}
The objectives are interpreted jointly: relative to $f_0$, we seek to improve AP under corruption and reduce latency and memory while limiting the loss of clean-data AP.
 
As shown in Fig.~\ref{fig:methodology_pipeline}, our framework addresses the two objectives through two coordinated components. For robustness, it pairs each clean frame with a corrupted counterpart, uses an ensemble reference to measure their prediction discrepancy, and orders the complete $2N$-sample training set from low to high perturbation sensitivity. For efficiency, it combines channel-width reduction with data-informed budget allocation. PointRCNN uses density-guided point- and neighbor-sampling budgets, whereas SECOND uses data-informed voxel-budget allocation. These architecture-specific budgets are determined offline. The resulting detector is trained from scratch on the ordered training set, with all construction decisions completed offline.

\subsection{Training Data Construction and Prioritization}
We prepare the training data by generating a corrupted counterpart for each clean frame, scoring each pair by perturbation sensitivity, and ordering all clean and corrupted samples from low to high sensitivity. The intuition is that this progression may help the detector learn stable geometric and object-level patterns from less corruption-sensitive frames before learning from more challenging ones.

\subsubsection{Corruption Injection}
\label{sec:corruption}

For each clean training sample $(\mathbf{x}_i,\mathbf{y}_i)\in\mathcal{D}$, we generate and cache one corrupted counterpart
$\tilde{\mathbf{x}}_i=C_{\tau_i}(\mathbf{x}_i)$, where $\tau_i$ denotes the corruption type assigned to sample $i$.
The corrupted point cloud retains the original annotation $\mathbf{y}_i$, forming the paired corrupted set
$\mathcal{D}^{c}=\{(\tilde{\mathbf{x}}_i,\mathbf{y}_i)\}_{i=1}^{N}$.
Each corrupted sample is generated once and reused for both sensitivity estimation and detector training, ensuring a consistent clean--corrupted pair throughout offline construction.

We consider six scene-level corruption types covering point-density changes, measurement noise, and adverse weather. Density decrease removes LiDAR points, whereas density increase introduces additional points. Gaussian noise perturbs point coordinates, and background noise injects randomly distributed clutter points. Rain and snow introduce precipitation-induced scattering and particle returns. All six corruption types are applied at severity level~4 of~5.

\subsubsection{Perturbation-Sensitivity Scoring}
\label{sec:sensitivity_prioritization}

For each clean--corrupted pair $(\mathbf{x}_i,\tilde{\mathbf{x}}_i)$, we quantify how much the detection results change after corruption. 
The detection results on the clean frame serve as a same-scene reference, separating corruption-induced changes from the scene's intrinsic difficulty. To reduce dependence on any single detector, we use $K=7$ pretrained fixed detectors including PointRCNN~\cite{shi2019pointrcnn}, PointRCNN-IoU~\cite{shi2019pointrcnn,li2020iou}, Part-$A^2$~\cite{shi2020parta2}, PointPillars~\cite{lang2019pointpillars}, SECOND~\cite{yan2018second}, SECOND-IoU~\cite{yan2018second,li2020iou}, and PV-RCNN~\cite{shi2020pvrcnn}, to produce ensemble-fused detections for both inputs. 
We then match the clean and corrupted detections by class and measure their confidence discrepancy, BEV overlap, and unmatched detections. We combine these discrepancies into a frame-level perturbation-sensitivity score $u_i$, where a larger value indicates that the frame is more strongly affected by corruption and should appear later in the prioritized training order for better learning.

We first establish correspondences between the ensemble-fused detections on the clean and corrupted point clouds. Within each object class, we greedily match boxes using a bird's-eye-view (BEV) intersection-over-union (IoU) threshold of $0.5$, with each box matched at most once. 
Let $\mathcal{M}_i$ denote the resulting set of matched box pairs for frame $i$. Matched pairs are used to measure confidence and BEV localization changes, while unmatched detections are accounted for separately.
For matched pairs $(a,b)\in\mathcal{M}_i$, we measure the absolute confidence difference and the discrepancy in BEV overlap. 
Averaging over the matched pairs gives
\begin{align}
\Delta_i^{\mathrm{score}}
&=
\frac{1}{|\mathcal{M}_i|}
\sum_{(a,b)\in\mathcal{M}_i}|s(a)-s(b)|,
\label{eq:score_diff}\\
\Delta_i^{\mathrm{box}}
&=
\frac{1}{|\mathcal{M}_i|}
\sum_{(a,b)\in\mathcal{M}_i}
\left[1-\operatorname{IoU}_{\mathrm{BEV}}(a,b)\right],
\label{eq:box_diff}
\end{align}
where $s(a)$ and $s(b)$ are the confidence scores of the clean and corrupted detections, respectively. The confidence discrepancy measures the magnitude of the change, regardless of whether confidence increases or decreases. Both terms are defined as zero when $\mathcal{M}_i$ is empty.

\begin{figure*}[!t]
  \centering
  \includegraphics[width=0.97\textwidth]{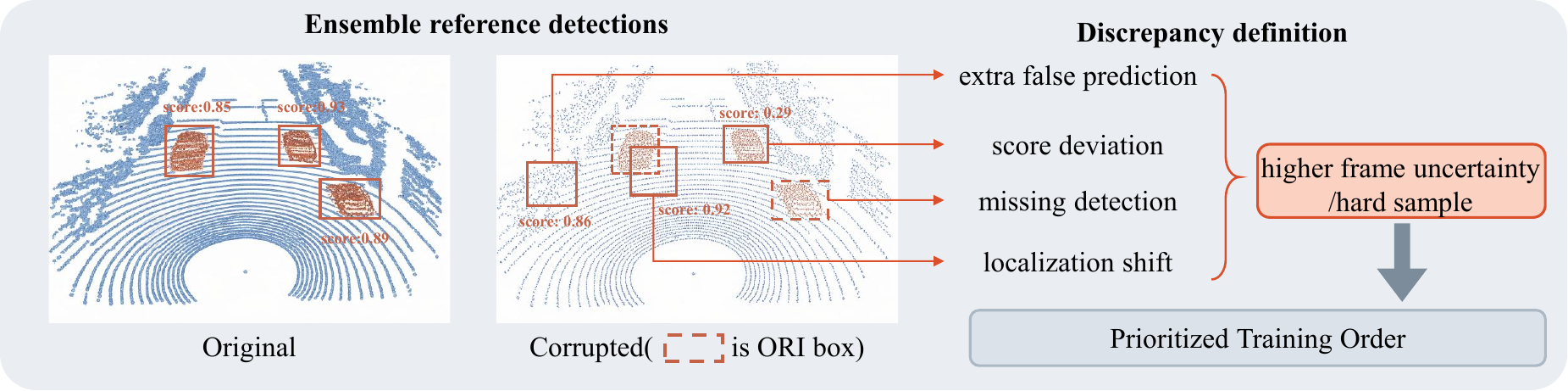}
  \caption{Case study of perturbation sensitivity. We compare ensemble predictions on an original point cloud and its corrupted version. Differences in false predictions, confidence scores, missing detections, and object locations are combined into a frame-level sensitivity score. A higher score indicates a harder sample that appears later in the prioritized training order.}
  \label{fig:uncertainty_case}
\end{figure*}

Detections without a match capture changes that the two matched-box terms do not cover. Let $n_{i,c}$ and $\tilde{n}_{i,c}$ denote the numbers of clean and corrupted detections for class $c$, respectively, and define
$Z_i=\sum_c\max(n_{i,c},\tilde{n}_{i,c})$.
We measure the unmatched-detection discrepancy as
\begin{equation}
\Delta_i^{\mathrm{miss}}=
\begin{cases}
1-|\mathcal{M}_i|/Z_i, & Z_i>0,\\
0, & Z_i=0.
\end{cases}
\label{eq:missing_rate}
\end{equation}
This term measures the deficit in matched detections relative to the sum of the larger detection count for each class. It captures unmatched predictions on either input. 
The frame-level perturbation-sensitivity score is then defined as
\begin{equation}
u_i =
\Delta_i^{\mathrm{score}}
+
\Delta_i^{\mathrm{box}}
+
\lambda_{\mathrm{miss}}\Delta_i^{\mathrm{miss}},
\qquad \lambda_{\mathrm{miss}}=0.5.
\label{eq:inconsistency}
\end{equation}
Fig.\ref{fig:uncertainty_case} intuitively illustrates the corruption-induced prediction discrepancies this score captures.

\subsubsection{Prioritized Training Order}
\label{sec:data_prioritization}

The perturbation-sensitivity score $u_i$ is computed for a clean--corrupted pair and assigned to both the clean frame $\mathbf{x}_i$ and its corrupted counterpart $\tilde{\mathbf{x}}_i$. The complete training list $\mathcal{S}$ contains all $N$ clean samples and their $N$ corrupted counterparts, each with its original annotation and assigned score. We obtain the prioritized list by sorting these $2N$ entries in ascending order of sensitivity:
\begin{equation}
\mathcal{S}_{\pi}
=
\operatorname{sort}_{\uparrow}(\mathcal{S};u).
\label{eq:prioritized_training_list}
\end{equation}
This ordering presents samples with smaller corruption-induced prediction changes before those with larger changes.

Prioritization changes only the training-sample presentation order: all samples are retained, and the detection loss and sample weights remain unchanged. To preserve this order during training, we disable shuffling and traverse the same prioritized training list $\mathcal{S}_{\pi}$ in every epoch.

\subsection{Resource-Efficient Detector Construction}
\label{sec:resource_effient}
While training data construction and prioritization target corruption robustness, we address deployment efficiency by reducing the detector's inference workload. We reduce channel widths in both PointRCNN and SECOND, representing point-based and voxel-based architectures, respectively. For PointRCNN, we additionally use offline point-density statistics to determine point and neighborhood sampling budgets within the network. For SECOND, offline point-count statistics additionally guide voxel-budget allocation. These modifications aim to reduce mean inference latency and peak allocated GPU memory. The channel widths and sampling budgets are specified before training and remain fixed throughout training and inference.

\subsubsection{Point-Based Detectors}

For PointRCNN, we reduce feature widths and sampling budgets for points and neighbors to decrease the inference workload. We scale the MLP and fully connected layer widths in the backbone, feature-propagation module, point head, and ROI head by $0.7$, rounding to implementation-compatible values. For example, the point-head and ROI-head widths change from $[256,256]$ to $[176,176]$.

Sampling budgets are determined from point-density statistics computed offline over the training set, after filtering points to the detector's input range. We measure the median valid-point count across training frames. and estimate the average neighbor count within each grouping radius using sampled query and reference points. These statistics guide the point and neighbor budgets, respectively.

For grouping branch $b$ at layer $l$, let $\bar{c}_{l,b}$ denote the average neighbor count and $S_{l,b}^{\mathrm{orig}}$ the original neighbor budget. We use their ratio to estimate local redundancy:
\begin{equation}
R_{l,b}
=
\max\left(
\frac{\bar{c}_{l,b}}{S_{l,b}^{\mathrm{orig}}},\,1
\right).
\label{eq:sensitivity_redundancy}
\end{equation}
The retention ratio for the neighbor-sampling budget and the corresponding integer-valued budget are given by
\begin{equation}
\begin{aligned}
\gamma_{l,b}
&=
\min\left(
\gamma_{\mathrm{budget}},
\max\left(\gamma_{\min},R_{l,b}^{-1/2}\right)
\right),\\
S_{l,b}^{\mathrm{new}}
&=
\max\left(
S_{\min},
\operatorname{round}
\left(\gamma_{l,b}S_{l,b}^{\mathrm{orig}}\right)
\right),
\end{aligned}
\label{eq:budget}
\end{equation}
where $0<\gamma_{\min}\leq\gamma_{\mathrm{budget}}\leq1$ bound the retained fraction, and $S_{\min}\leq S_{l,b}^{\mathrm{orig}}$ prevents excessively small neighbor-sampling budget. Higher redundancy therefore permits a greater budget reduction, subject to these bounds.

For point sampling, let $\tilde{N}$ be the median valid-point count and $N_{\mathrm{ref}}$ the reference input size. We compute a shared scaling ratio and apply it to each layer:
\begin{equation}
\begin{aligned}
\gamma^{\mathrm{point}}
&=
\min\left(
\gamma_{\mathrm{budget}}^{\mathrm{point}},
\max\left(
\gamma_{\min}^{\mathrm{point}},
\sqrt{\frac{\tilde{N}}{N_{\mathrm{ref}}}}
\right)
\right),\\
N_l^{\mathrm{new}}
&=
\max\left(
1,
\operatorname{round}
\left(\gamma^{\mathrm{point}}N_l^{\mathrm{orig}}\right)
\right),
\end{aligned}
\label{eq:npoint_ratio}
\end{equation}
where $0<\gamma_{\min}^{\mathrm{point}}
\leq\gamma_{\mathrm{budget}}^{\mathrm{point}}\leq1$.
The same budget rules apply to applicable ROI point sampling and neighborhood aggregation; stages that aggregate all points retain that setting. The resulting network widths and sampling budgets remain fixed throughout training and inference.

\subsubsection{Voxel-Based Detectors}

SECOND groups points into voxels and extracts features through sparse 3D convolutions followed by a BEV backbone.
For SECOND, we combine channel-width reduction in the BEV backbone and upsampling layers with data-informed voxel-budget allocation. Offline data statistics guide the voxel budget used by the detector. This voxel-based allocation complements the point- and neighbor-sampling budget allocation used for PointRCNN. The resulting detector is trained from scratch on the prioritized training list $\mathcal{S}_{\pi}$ using the constructed channel-width and voxel-budget configuration.

\subsection{Complete Procedure}
\label{sec:algorithm}
Both detectors are trained from scratch on the prioritized training list $\mathcal{S}_{\pi}$, with the reduced channel widths and architecture-specific budgets fixed throughout training and inference. Algorithm~\ref{alg:pipeline} jointly targets corruption robustness and deployment efficiency. Given $\mathcal{D}$, $C$, $\{f_k\}_{k=1}^{K}$, $\mathcal{A}$, construction settings, and $T$, it initializes the training list (line~1), generates corrupted counterparts (lines~2--3), fuses reference predictions (line~4), matches detections (line~5), and computes perturbation-sensitivity scores (line~6). Each pair is added with its annotation and shared score (line~7), and the complete list is sorted into $\mathcal{S}_{\pi}$ (line~9).

For efficiency, the algorithm reduces feature widths (line~10) and, where applicable, uses offline density statistics to allocate sampling budgets (lines~11--13); otherwise, spatial-processing settings are retained (lines~14--16). It then instantiates the reduced detector (line~17), trains it from scratch on $\mathcal{S}_{\pi}$ for $T$ epochs (line~18), and returns $f_{\theta'}$ (line~19).

\begin{algorithm}[!t]
\footnotesize
\caption{Offline Robust and Resource-Efficient Detector Construction}
\label{alg:pipeline}
\begin{algorithmic}[1]
\REQUIRE Clean set $\mathcal{D}$; corruption operator $C$;
reference detectors $\{f_k\}_{k=1}^{K}$; architecture $\mathcal{A}$;
construction settings; epochs $T$
\ENSURE Trained detector $f_{\theta'}$

\STATE Initialize an empty list $\mathcal{S}$
\FOR{$i=1$ \TO $N$}
    \STATE Generate and cache
    $\tilde{\mathbf{x}}_i \leftarrow C_{\tau_i}(\mathbf{x}_i)$
    \STATE Fuse reference detections separately on
    $\mathbf{x}_i$ and $\tilde{\mathbf{x}}_i$
    \STATE Match detections by class using BEV IoU $\geq 0.5$
    \STATE Compute $u_i$ using \eqref{eq:inconsistency}
    \STATE Append $(\mathbf{x}_i,\mathbf{y}_i,u_i)$ and
    $(\tilde{\mathbf{x}}_i,\mathbf{y}_i,u_i)$ to $\mathcal{S}$
\ENDFOR
\STATE $\mathcal{S}_{\pi}
\leftarrow \operatorname{sort}_{\uparrow}(\mathcal{S};u)$

\STATE Reduce architecture-specific feature widths

\COMMENT{point-processing or BEV layers}
\IF{$\mathcal{A}$ uses point sampling and neighborhood aggregation}
    \STATE Estimate $\tilde{N}$ and $\bar{c}_{l,b}$ over $\mathcal{D}$
    \STATE Set $S_{l,b}^{\mathrm{new}}$ and $N_l^{\mathrm{new}}$
    using \eqref{eq:budget} and \eqref{eq:npoint_ratio}

\ENDIF

\STATE Instantiate $f_{\theta'}$ with the resulting settings
\STATE Train from scratch on $\mathcal{S}_{\pi}$ for $T$ epochs,
preserving its order in every epoch
\RETURN $f_{\theta'}$
\end{algorithmic}
\end{algorithm}

\section{Evaluation}
\label{sec:eval}

We evaluate our offline provisioning framework against its two objectives: robustness to point-cloud corruption and deployment efficiency. We also examine changes in clean-data accuracy and the contributions of the framework components. Our experiments address three research questions:
\begin{itemize}
    \item \textbf{RQ1: Corruption Robustness.}
    How does our framework affect detection accuracy under point-cloud corruption and on clean data?
    \item \textbf{RQ2: Deployment Efficiency.}
    How much does our framework reduce mean inference latency and peak allocated GPU memory compared with the original detectors and baseline methods?
    \item \textbf{RQ3: Component Contributions.}
    How do perturbation-sensitivity-based data prioritization and resource-efficient detector construction contribute to the observed results?
\end{itemize}

\subsection{Experimental Setup}
\label{sec:setup}

We evaluate our framework on KITTI~\cite{geiger2012kitti} and ONCE~\cite{mao2021once} using PointRCNN~\cite{shi2019pointrcnn} and SECOND~\cite{yan2018second}, representing point-based and voxel-based detector architectures, respectively. For both detectors, we evaluate Car, Pedestrian, and Cyclist on the clean validation sets and their corrupted counterparts.

Experiments use OpenPCDet~v0.6~\cite{openpcdet2020}, PyTorch~1.13, CUDA~11.3, and one NVIDIA RTX~4090D.

We evaluate detection accuracy using average precision (AP) on KITTI and ONCE~\cite{geiger2012kitti,mao2021once}. Following previous work~\cite{dong2023benchmarking,li2023gpa3d}, KITTI results use AP$_{\mathrm{R40}}$, evaluated at 40 recall positions, and are reported at the Easy, Moderate, and Hard difficulty levels. For ONCE, we also compute AP at 40 recall positions, using an R40 variant of the orientation-aware evaluation procedure~\cite{mao2021once}, and report results across the evaluated distance intervals. Both datasets therefore use the same recall sampling resolution, while retaining their dataset-specific matching and evaluation criteria.

\subsection{RQ1: Corruption Robustness}
\label{sec:rq1}

We evaluate corruption robustness using average precision (AP) under corruption and report clean AP to assess the accompanying changes in clean-data accuracy. Using the dataset-specific protocols described in Section~\ref{sec:setup}, we use KITTI Moderate AP for the main comparisons and report ONCE AP over three distance intervals: 0--30~m, 30--50~m, and beyond 50~m. For compact comparisons, we additionally report the arithmetic mean of the three ONCE distance-bin AP values, termed range-average AP. 

Each clean frame has one fixed corrupted counterpart, generated with the toolkit of~\cite{robustness_pc_detector} using the six corruption types in Section~\ref{sec:corruption} at severity level~4 of~5. For the proposed method, the training set contains all clean training frames and their cached corrupted counterparts in a 1:1 ratio. AP under corruption is evaluated in one pass over the mixed corrupted validation set, rather than averaged over six separate per-type evaluations. The results therefore characterize the evaluated mixed-corruption protocol.
The complete baseline accuracy comparison focuses on Car, while the cross-class comparison includes Original, Network Slimming, and Ours.
\textbf{Original} uses the released checkpoint and default detector configuration. \textbf{Retrain} is the full-size detector retrained on the 1:1 clean--corrupted mixture without channel-width or architecture-specific budget reallocation. \textbf{GMP} applies unstructured magnitude pruning, \textbf{LTH/IMP} applies iterative magnitude pruning with weight rewinding, and \textbf{Network Slimming} (Net.\ Slim.\ in the tables) reduces channels using batch-normalization scales. 
\textbf{Ours} combines perturbation-sensitivity-based training-data prioritization with resource-efficient detector construction. For PointRCNN, it uses channel-width reduction and density-guided point- and neighbor-sampling budgets. For SECOND, it uses channel-width reduction in the BEV backbone and upsampling layers together with data-informed voxel-budget allocation.

Table~\ref{tab:cross_class} shows that AP under corruption improves over Original in all six detector--class combinations on KITTI and five of six on ONCE. The mean gains are 3.26 points in KITTI Moderate AP and 4.80 points in ONCE range-average AP. Across the twelve dataset--detector--class combinations, the descriptive mean gain is 4.03 points. The exception is PointRCNN Pedestrian on ONCE, whose AP under corruption decreases from 4.19 to 3.87, a loss of 0.32 points. A plausible explanation for the overall gains is the combination of corrupted-data training and perturbation-sensitivity ordering: paired clean and corrupted frames expose the detector to the evaluated degradations, while the low-to-high-sensitivity order is intended to support progressive learning. RQ3 examines the contributions of the training order and ensemble reference.

\begin{table}[!htbp]
\centering
\caption{Detection Accuracy across Datasets, Detectors, and Classes}
\label{tab:cross_class}
\begin{threeparttable}
{\footnotesize
\setlength{\tabcolsep}{2.6pt}
\renewcommand{\arraystretch}{1.08}
\begin{tabularx}{\columnwidth}{@{}l>{\raggedright\arraybackslash}Xcccc@{}}
\toprule
\textbf{Class} & \textbf{Method} &
\multicolumn{2}{c}{\shortstack{\textbf{KITTI}\\\textbf{Moderate}}} &
\multicolumn{2}{c}{\shortstack{\textbf{ONCE}\\\textbf{Range Avg.}}}\\
\cmidrule(lr){3-4}\cmidrule(lr){5-6}
&& \textbf{Clean} & \textbf{Corr.} & \textbf{Clean} & \textbf{Corr.}\\
\midrule
\multicolumn{6}{@{}l}{\textbf{PointRCNN}}\\
\midrule
\multirow{3}{*}{Car} & Original & 80.65 & 61.03 & 43.78 & 25.02 \\
 & Net.\ Slim. & 77.79 & 56.78 & 42.95 & 30.37 \\
 & \textbf{Ours} & 80.04 & 64.35 & 43.47 & 32.20 \\
\cmidrule(lr){1-6}
\multirow{3}{*}{Pedestrian} & Original & 54.81 & 43.86 & 4.95 & 4.19 \\
 & Net.\ Slim. & 57.28 & 45.10 & 2.40 & 2.39 \\
 & \textbf{Ours} & 57.36 & 46.26 & 3.98 & 3.87 \\
\cmidrule(lr){1-6}
\multirow{3}{*}{Cyclist} & Original & 71.09 & 47.94 & 26.41 & 16.45 \\
 & Net.\ Slim. & 68.40 & 49.00 & 20.79 & 16.34 \\
 & \textbf{Ours} & 69.50 & 52.81 & 23.95 & 19.90 \\
\cmidrule(lr){1-6}
\multicolumn{6}{@{}l}{\textbf{SECOND}}\\
\midrule
\multirow{3}{*}{Car} & Original & 81.61 & 62.17 & 68.74 & 35.92 \\
 & Net.\ Slim. & 78.84 & 66.19 & 67.84 & 45.31 \\
 & \textbf{Ours} & 79.05 & 67.10 & 69.86 & 46.72 \\
\cmidrule(lr){1-6}
\multirow{3}{*}{Pedestrian} & Original & 51.15 & 36.89 & 27.64 & 16.64 \\
 & Net.\ Slim. & 43.75 & 36.60 & 26.26 & 17.55 \\
 & \textbf{Ours} & 45.01 & 37.70 & 28.92 & 18.46 \\
\cmidrule(lr){1-6}
\multirow{3}{*}{Cyclist} & Original & 66.74 & 45.25 & 52.50 & 29.62 \\
 & Net.\ Slim. & 44.75 & 47.25 & 51.26 & 33.82 \\
 & \textbf{Ours} & 66.05 & 48.50 & 53.87 & 35.50 \\
\bottomrule
\end{tabularx}}
\begin{tablenotes}[flushleft]
\footnotesize
\item KITTI values are Moderate AP$_{\mathrm{R40}}$. ONCE values are unweighted arithmetic means of the AP values in the 0--30~m, 30--50~m, and beyond-50~m bins. Clean and Corr.\ denote evaluation on clean and corrupted validation inputs, respectively.
\end{tablenotes}
\end{threeparttable}
\end{table}

\begin{table*}[!tb]
\centering
\caption{Car Detection Accuracy on Clean and Corrupted Data}
\label{tab:car_accuracy}
\begin{threeparttable}
{\footnotesize
\setlength{\tabcolsep}{2.6pt}
\renewcommand{\arraystretch}{1.08}
\begin{tabular}{@{}llcccccccccccc@{}}
\toprule
\multicolumn{2}{c}{} & \multicolumn{6}{c}{\textbf{KITTI}} & \multicolumn{6}{c}{\textbf{ONCE}}\\
\cmidrule(lr){3-8}\cmidrule(lr){9-14}
\textbf{Detector} & \textbf{Method} &
\multicolumn{3}{c}{\textbf{Clean}} & \multicolumn{3}{c}{\textbf{Corrupted}} &
\multicolumn{3}{c}{\textbf{Clean}} & \multicolumn{3}{c}{\textbf{Corrupted}}\\
\cmidrule(lr){3-5}\cmidrule(lr){6-8}\cmidrule(lr){9-11}\cmidrule(lr){12-14}
& & Easy & Moderate & Hard & Easy & Moderate & Hard & 0--30m & 30--50m & $>$50m & 0--30m & 30--50m & $>$50m\\
\midrule
\multirow{6}{*}{\textbf{PointRCNN}} & Original & 91.62 & 80.65 & 78.07 & 77.08 & 61.03 & 56.90 & 73.83 & 39.83 & 17.68 & 49.60 & 17.85 & 7.62 \\
 & Retrain & 88.71 & 80.01 & 77.35 & 82.13 & 64.73 & 60.08 & 73.21 & 39.80 & 17.44 & 64.55 & 24.31 & 7.74 \\
 & GMP & 74.02 & 62.84 & 60.37 & 62.53 & 49.16 & 43.84 & 57.34 & 27.63 & 12.65 & 45.92 & 16.46 & 5.52 \\
 & LTH/IMP & 89.11 & 77.85 & 75.15 & 79.56 & 62.31 & 57.72 & 70.96 & 37.13 & 14.62 & 60.02 & 21.83 & 5.31 \\
 & Net.\ Slim. & 89.29 & 77.79 & 73.14 & 73.52 & 56.78 & 50.59 & 72.95 & 39.36 & 16.55 & 61.88 & 22.03 & 7.20 \\
 & \textbf{Ours} & 89.56 & 80.04 & 77.53 & 81.71 & 64.35 & 60.01 & 73.05 & 39.86 & 17.50 & 64.57 & 24.40 & 7.64 \\
\midrule
\multirow{6}{*}{\textbf{SECOND}} & Original & 90.55 & 81.61 & 78.60 & 77.18 & 62.17 & 58.68 & 85.21 & 66.99 & 54.02 & 55.25 & 31.33 & 21.18 \\
 & Retrain & 89.23 & 78.02 & 75.30 & 83.13 & 67.84 & 64.49 & 85.00 & 68.82 & 55.39 & 72.28 & 41.09 & 25.59 \\
 & GMP & 88.75 & 78.97 & 75.85 & 81.57 & 65.32 & 61.86 & 80.33 & 57.05 & 42.77 & 65.97 & 33.88 & 19.47 \\
 & LTH/IMP & 88.88 & 78.73 & 75.52 & 81.22 & 65.05 & 61.63 & 83.14 & 63.44 & 50.01 & 69.62 & 37.95 & 22.71 \\
 & Net.\ Slim. & 88.18 & 78.84 & 75.79 & 81.54 & 66.19 & 62.52 & 83.49 & 66.42 & 53.62 & 71.70 & 39.99 & 24.24 \\
 & \textbf{Ours} & 89.49 & 79.05 & 75.94 & 82.71 & 67.10 & 62.64 & 86.20 & 68.97 & 54.41 & 72.51 & 41.88 & 25.77 \\
\bottomrule
\end{tabular}}
\begin{tablenotes}[flushleft]
\footnotesize
\item KITTI reports Easy/Moderate/Hard AP$_{\mathrm{R40}}$; ONCE reports range-wise AP. Baselines are not budget-matched.
\end{tablenotes}
\end{threeparttable}
\end{table*}

Table~\ref{tab:car_accuracy} compares all six methods on Car detection. On KITTI, Ours raises Moderate AP under corruption from 61.03 to 64.35 for PointRCNN and from 62.17 to 67.10 for SECOND, compared with the corresponding Original baselines. Relative to Network Slimming, the gains are 7.57 and 0.91 points, respectively. Retrain achieves slightly higher values of 64.73 and 67.84. On ONCE, Ours and Retrain both attain 32.20 range-average AP under corruption with PointRCNN after rounding, while Ours reaches 46.72 with SECOND compared with 46.32 for Retrain. Across these four dataset--detector comparisons, the reported AP-under-corruption summaries for Ours differ from those for full-size Retrain by at most 0.74 points. RQ2 reports the corresponding latency and memory measurements. These comparisons reflect the complete configurations and do not isolate the effects of training sample order, channel widths, or architecture-specific budgets.

On KITTI, PointRCNN Car loses 0.61 points of clean Moderate AP while gaining 3.32 points under corruption. SECOND Pedestrian presents a larger trade-off: clean AP decreases from 51.15 to 45.01, while AP under corruption increases from 36.89 to 37.70. On ONCE, PointRCNN loses clean range-average AP in all three classes, by 0.31, 0.97, and 2.46 points, whereas SECOND gains 1.12, 1.28, and 1.37 points. PointRCNN Pedestrian on ONCE loses both clean AP and AP under corruption. These differences show that the benefits are not uniform across architectures and classes. The observed accuracy trade-offs reflect the performance of the complete configurations, which jointly incorporate training-data prioritization, channel-width reduction, and architecture-specific budget allocation. The results do not separate the effects of these possible effects, but show why clean AP must be considered alongside AP under corruption.

For ONCE Car detection, PointRCNN improves AP under corruption by 14.97, 6.55, and 0.02 points in the 0--30~m, 30--50~m, and beyond-50~m intervals, respectively. SECOND improves by 17.26, 10.55, and 4.59 points. The PointRCNN gain is concentrated in the nearer intervals, whereas SECOND improves in all three. Perturbation-sensitivity ordering changes how training samples are presented, but does not restore missing measurements at inference; this may limit its benefits when the available geometric information is insufficient. The near-zero PointRCNN gain beyond 50~m also indicates that the aggregate improvement should not be interpreted as a uniform long-range benefit. 

\subsection{RQ2: Deployment Efficiency}
\label{sec:rq2}

We measure deployment efficiency using mean inference latency and peak allocated GPU memory, and report the 95th-percentile (p95) latency to characterize slower inference iterations. Measurements use batch size~1 on the hardware in Section~\ref{sec:setup}. Each checkpoint undergoes 100 warm-up iterations followed by 500 timed iterations, with CUDA synchronization on every iteration. We repeat the measurement five times for each clean and corrupted input condition. Latency is reported in milliseconds; peak allocated GPU memory is measured in MB with \texttt{torch.cuda.max\_memory\_allocated()}.

Table~\ref{tab:runtime_profiles} shows that Ours reduces mean inference latency and peak allocated GPU memory relative to Original in all eight settings covering two datasets, two detectors, and clean and corrupted inputs. Averaging the relative reductions across these settings gives 14.4\% lower mean latency and 38.0\% lower peak allocated GPU memory. For clean KITTI inputs, PointRCNN latency decreases from 47.96 to 38.46~ms (19.8\%), and memory from 408.1 to 188.2~MB (53.9\%). On clean ONCE inputs, the corresponding reductions are 25.3\% and 53.7\%. SECOND achieves smaller reductions: 7.8\% in latency and 22.5\% in memory on clean KITTI inputs, and 2.6\% and 22.7\% on clean ONCE inputs. Both detectors combine channel-width reduction with data-informed budget allocation, but the budgeted operations differ: PointRCNN adjusts point and neighbor sampling, whereas SECOND adjusts the voxel budget. The reported measurements capture the combined effects of these modifications and do not isolate their individual contributions to the resource savings.

\begin{table*}[!tb]
\centering
\caption{Inference Latency and Peak Allocated GPU Memory on the RTX~4090D}
\label{tab:runtime_profiles}
\begin{threeparttable}
{\footnotesize
\setlength{\tabcolsep}{2.6pt}
\renewcommand{\arraystretch}{1.08}
\begin{tabular}{@{}llcccccc@{}}
\toprule
&& \multicolumn{3}{c}{\textbf{KITTI}} & \multicolumn{3}{c}{\textbf{ONCE}}\\
\cmidrule(lr){3-5}\cmidrule(lr){6-8}
\textbf{Detector} & \textbf{Method} &
\shortstack{\textbf{Mean}\\\textbf{latency}} &
\shortstack{\textbf{p95}\\\textbf{latency}} &
\shortstack{\textbf{Peak}\\\textbf{memory}} &
\shortstack{\textbf{Mean}\\\textbf{latency}} &
\shortstack{\textbf{p95}\\\textbf{latency}} &
\shortstack{\textbf{Peak}\\\textbf{memory}}\\
\midrule
\multirow{6}{*}{\textbf{PointRCNN}} & Original & 47.96/48.88 & 49.27/50.79 & 408.1/407.7 & 669.30/670.48 & 677.65/695.51 & 633.3/633.3 \\
 & Retrain & 46.92/46.91 & 48.71/48.74 & 408.1/407.7 & 669.17/670.52 & 677.04/696.29 & 633.3/633.3 \\
 & GMP & 47.03/47.65 & 48.74/49.97 & 440.4/440.4 & 669.92/670.40 & 677.27/687.38 & 666.0/666.0 \\
 & LTH/IMP & 47.50/47.59 & 50.03/49.77 & 440.4/440.4 & 669.80/670.23 & 677.86/678.99 & 666.0/666.0 \\
 & Net.\ Slim. & 45.55/45.82 & 47.56/48.34 & 331.4/331.4 & 666.93/666.26 & 675.05/673.82 & 468.8/468.8 \\
 & \textbf{Ours} & 38.46/38.46 & 40.52/40.42 & 188.2/188.2 & 499.99/499.74 & 507.20/506.53 & 293.5/293.5 \\
\midrule
\multirow{6}{*}{\textbf{SECOND}} & Original & 12.74/12.65 & 13.38/12.94 & 292.1/297.3 & 16.52/16.73 & 19.77/20.31 & 355.1/367.8 \\
 & Retrain & 12.73/12.75 & 13.09/13.06 & 292.1/297.3 & 16.53/16.44 & 20.18/19.92 & 355.1/367.8 \\
 & GMP & 13.32/13.31 & 13.65/13.76 & 332.4/339.9 & 16.97/17.12 & 20.86/21.15 & 396.4/410.9 \\
 & LTH/IMP & 13.10/13.17 & 14.28/14.36 & 327.0/334.8 & 16.74/16.72 & 20.55/21.04 & 391.0/405.5 \\
 & Net.\ Slim. & 12.26/12.09 & 12.76/12.58 & 230.2/238.1 & 16.15/15.99 & 19.80/19.66 & 287.9/302.9 \\
 & \textbf{Ours} & 11.75/11.73 & 12.17/12.21 & 226.3/232.4 & 16.09/15.76 & 19.84/19.42 & 274.4/287.8 \\
\bottomrule
\end{tabular}}
\begin{tablenotes}[flushleft]
\footnotesize
\item Each entry is clean/corrupted. Mean and p95 are milliseconds; memory is peak allocated GPU memory in MB.
\end{tablenotes}
\end{threeparttable}
\end{table*}

Ours also achieves lower mean latency and peak allocated GPU memory than the evaluated baselines. On clean KITTI inputs, PointRCNN Retrain uses 46.92~ms and 408.1~MB, remaining close to Original. Network Slimming uses 45.55~ms and 331.4~MB; Ours reduces these values by 15.6\% and 43.2\%, respectively, while improving Moderate AP under corruption by 7.57 points (RQ1). GMP and LTH/IMP retain similar latencies of 47.03 and 47.50~ms and increase peak allocated GPU memory to 440.4~MB. These results are consistent with reducing the dimensions and sampling workload of the executed network: in the evaluated implementations, Ours achieves larger resource savings than GMP and LTH/IMP. The comparisons concern the reported configurations, not methods matched to a common resource budget.

The mean-latency reductions under corrupted inputs are 21.3\% and 25.5\% for PointRCNN on KITTI and ONCE, and 7.3\% and 5.8\% for SECOND. Relative to Original, the reported p95 latency is lower in seven of the eight evaluated settings. The only exception is SECOND on clean ONCE inputs, where p95 latency increases marginally from 19.77 to 19.84 ms, a difference of 0.07 ms. Overall, the results show lower p95 latency across the evaluated settings. These measurements demonstrate resource savings on GPU, and indicate the framework's potential to facilitate real-time detection on resource-constrained onboard hardware.

\subsection{RQ3: Component Contributions}
\label{sec:rq3}

We evaluate component contributions on KITTI using PointRCNN, with clean AP and AP under corruption as the evaluation metrics. The training-order comparison uses the full-size detector and the 1:1 clean--corrupted mixture for Car detection. The component ablation study compares the full resource-efficient detector with variants without ensemble reference, data prioritization, or density-guided sampling across Car, Pedestrian, and Cyclist. We report Moderate AP in the text and all three difficulty levels in the tables.

Table~\ref{tab:data_selection} shows that perturbation-sensitivity ordering achieves the highest AP across all six clean and corrupted results. Compared with random order, clean Moderate AP increases from 79.59 to 82.12, and Moderate AP under corruption from 64.86 to 67.12, gains of 2.53 and 2.26 points. The corresponding gains over confidence-based ordering are 1.46 and 0.54 points. In the reported run, low-to-high perturbation-sensitivity ordering achieves higher AP than the two alternative ordering strategies. This pattern supports the use of corruption-induced prediction changes to determine the training order and are consistent with the intended progression from easier to more challenging samples.

\begin{table}[!htbp]
\centering

\caption{Effect of Training Order on KITTI Car with Full-Size PointRCNN (AP$_{\mathrm{R40}}$, IoU 0.7)}
\label{tab:data_selection}
\vspace{-5pt}
{\footnotesize
\setlength{\tabcolsep}{2.6pt}
\renewcommand{\arraystretch}{1.08}
\begin{tabularx}{\columnwidth}{@{}>{\raggedright\arraybackslash}Xcccccc@{}}
\toprule
& \multicolumn{3}{c}{\textbf{Clean}} & \multicolumn{3}{c}{\textbf{Corrupted}} \\
\cmidrule(lr){2-4}\cmidrule(lr){5-7}
\textbf{Training order} & E & M & H & E & M & H \\
\midrule
Random order & 88.05 & 79.59 & 77.06 & 82.07 & 64.86 & 61.98 \\
Confidence-based & 89.32 & 80.66 & 77.96 & 81.86 & 66.58 & 62.20 \\
Perturbation-sensitivity & \textbf{91.27} & \textbf{82.12} & \textbf{79.50} & \textbf{84.42} & \textbf{67.12} & \textbf{62.52} \\
\bottomrule
\end{tabularx}}

\vspace{8pt}

\caption{Component Ablation on KITTI with PointRCNN}
\label{tab:ablation}
\vspace{-5pt}
\begin{threeparttable}
{\footnotesize
\setlength{\tabcolsep}{2.6pt}
\renewcommand{\arraystretch}{1.08}
\begin{tabularx}{\columnwidth}{@{}>{\raggedright\arraybackslash}Xcccccc@{}}
\toprule
\textbf{Variant} & \multicolumn{3}{c}{\textbf{Clean}} & \multicolumn{3}{c}{\textbf{Corrupted}}\\
\cmidrule(lr){2-4}\cmidrule(lr){5-7}
& E & M & H & E & M & H\\
\midrule
\multicolumn{7}{@{}l}{\textbf{Car}}\\
\midrule
\textbf{Full pipeline} & 89.56 & 80.04 & 77.53 & 81.71 & 64.35 & 60.01 \\
No ensemble reference & 87.49 & 78.89 & 74.91 & 80.13 & 63.20 & 57.28 \\
No data prioritization & 87.38 & 77.22 & 76.55 & 78.77 & 63.20 & 58.93 \\
No density-guided sampling & 86.71 & 76.95 & 75.25 & 80.45 & 62.88 & 55.33 \\
\midrule
\multicolumn{7}{@{}l}{\textbf{Pedestrian}}\\
\midrule
\textbf{Full pipeline} & 64.51 & 57.36 & 48.96 & 53.74 & 46.26 & 39.72 \\
No ensemble reference & 53.58 & 45.65 & 39.46 & 44.93 & 37.38 & 32.74 \\
No data prioritization & 56.48 & 48.18 & 42.61 & 47.10 & 39.22 & 34.57 \\
No density-guided sampling & 63.32 & 54.91 & 48.14 & 52.24 & 40.97 & 36.81 \\
\midrule
\multicolumn{7}{@{}l}{\textbf{Cyclist}}\\
\midrule
\textbf{Full pipeline} & 89.01 & 69.50 & 65.84 & 76.07 & 52.81 & 48.88 \\
No ensemble reference & 85.58 & 66.83 & 62.40 & 72.57 & 48.30 & 45.55 \\
No data prioritization & 85.37 & 67.28 & 63.81 & 73.55 & 49.97 & 47.11 \\
No density-guided sampling & 86.43 & 67.87 & 63.41 & 71.40 & 48.14 & 45.75 \\
\bottomrule
\end{tabularx}}
\begin{tablenotes}[flushleft]
\footnotesize
\item AP$_{\mathrm{R40}}$ on clean data and under corruption. Results compare the full pipeline with the reported component variants.
\end{tablenotes}
\end{threeparttable}

\end{table}

Table~\ref{tab:ablation} shows lower Moderate AP under corruption for the \emph{No ensemble reference} variant than for the full pipeline, by 1.15, 8.88, and 4.51 points for Car, Pedestrian, and Cyclist, respectively. The \emph{No data prioritization} variant loses 1.15, 7.04, and 2.84 points. Both variants also have lower clean Moderate AP in all three classes. These results support the roles of the ensemble reference and data prioritization in robust training, with the largest observed differences for Pedestrian. The ensemble reference is intended to reduce dependence on any single detector when ordering samples, although these results do not establish why the gains vary by class.

The \emph{No density-guided sampling} variant achieves 62.88, 40.97, and 48.14 Moderate AP under corruption for Car, Pedestrian, and Cyclist, respectively, compared with 64.35, 46.26, and 52.81 for the full pipeline. Relative to this variant, the full pipeline improves Moderate AP by 1.47, 5.29, and 4.67 points under corruption and by 3.09, 2.45, and 1.63 points on clean data, respectively. These results are consistent with the intended role of density-guided allocation in retaining useful geometric information during sampling reduction. The improvements across all three classes under both clean and corrupted conditions provide empirical support for incorporating density-guided sampling-budget allocation into the resource-efficient detector construction process.

\FloatBarrier

\section{Discussion and Limitations}
\label{sec:discussion}

Our results suggest that corruption robustness and deployment efficiency can improve together through offline detector provisioning. Perturbation-sensitivity-based data prioritization guides robust training, while channel-width reduction and data-informed budget allocation target reduce inference workload. However, clean AP does not improve uniformly across detectors and classes. Deployment decisions should therefore consider clean AP and AP under corruption alongside mean inference latency and peak allocated GPU memory, rather than rely on any one metric.

Offline construction adds the cost of ensemble inference, sensitivity scoring, and density analysis to detector training. These operations are performed before deployment and do not enter the inference path of the constructed detector. This separates the cost of preparing training data and determining network widths and architecture-specific sampling budgets from the cost of processing each incoming frame. We have not quantified the total offline cost, so the reported inference savings alone do not quantify the overall computational savings across training and deployment.

The evaluation covers two datasets, two detector architectures, six synthetic corruption types at one severity level, and one desktop GPU. Generalization to real adverse weather, other architectures, and onboard hardware remains untested. Accuracy results are averaged over multiple training runs. The baselines are not matched to a common resource budget; small AP differences and the effects of individual reduction mechanisms therefore require further validation. AP measures detection quality rather than downstream driving safety, and peak allocated GPU memory excludes memory outside the measured PyTorch allocations. Matched-budget comparisons, and evaluation under real corruptions on onboard hardware would strengthen the evidence.

\section{Conclusion}
\label{sec:conclusion}
We studied the offline provisioning of robust and resource-efficient LiDAR perception services for autonomous driving, where a provider constructs a fixed detector before deployment so that no adaptation is needed while serving inference requests. 
The proposed framework orders the complete clean and corrupted training set from least to most perturbation-sensitive, estimates sensitivity using an ensemble of seven reference detectors, and reduces inference cost by narrowing channel widths and allocating architecture-specific sampling budgets based on offline data statistics. Across KITTI and ONCE, two detector architectures, and three object classes, the provisioned detectors improve AP under corruption by 4.03 points on average over the original detectors while reducing mean inference latency and peak allocated GPU memory by 14.4\% and 38.0\%, with AP under corruption improving in eleven of the twelve evaluated combinations and configuration-dependent trade-offs in clean-data AP. For Car detection on KITTI with PointRCNN, for example, AP under corruption increases by 3.32 points while mean latency and peak memory decrease by 19.8\% and 53.9\%. These results show that corruption robustness and deployment efficiency need not be traded against each other when both are addressed in one offline construction process. Future work includes comparing against baselines under matched resource budgets, measuring latency and memory on onboard hardware, and reporting per-corruption results over multiple training seeds.

\section*{Data Availability}

KITTI and ONCE are publicly available, and we generate corruptions as in~\cite {robustness_pc_detector}.
Our code is available at \url{https://github.com/drunklibai/robust-efficient-lidar-perception}.

\section*{Acknowledgement}
We thank the editors and anonymous reviewers for their time and comments. This research received partial support from the National Natural Science Foundation of China (Grants No. U24A20337, 62372228), Shenzhen-Hong Kong-Macau Technology Research Programme (Type C) (Grant No. SGDX20230821091559018).
The opinions, findings, conclusions, or recommendations expressed in this work are solely those of the author(s) and do not necessarily reflect the views of the National Research Foundation.

\balance

\bibliographystyle{IEEEtranN}
\bibliography{references}

\end{document}